\documentclass[12pt]{article}

\usepackage{amssymb,amsmath,amsfonts,latexsym,graphicx}
\usepackage[numbers]{natbib}
\usepackage{multirow}
\usepackage{hyperref} 
\hypersetup{
    colorlinks=true,
    urlcolor=blue,
    citecolor=black,
    pdftitle={Overleaf Example},
    pdfpagemode=FullScreen,
}

\begin{document}

\begin{center}
\Large \bf GALAR-TemporalNet v2: Anatomy-Guided Dual-Branch Temporal Classification with Bidirectional Mamba and Dual-Graph GCN for Video Capsule Endoscopy---after competition results \rm







\vspace{1cm}


\large Jiye Won$^{a}$, Seangmin Lee$^{a}$, Soon Ki Jung$^{a,*}$


\vspace{0.5cm}

\normalsize


$^{a}$ School of Computer Science and Engineering, Kyungpook National University, Daegu, Republic of Korea\\
$^{*}$ Corresponding author

\vspace{5mm}


Corresponding Author Email: {\tt skjung@knu.ac.kr} 

Team Name: {\tt Dubai Chewy Cookie}

Updated GitHub Repository Link: {\url{https://github.com/blueberry0814/GALAR-TemporalNet_v2.git}}

\vspace{1cm}

\end{center}

\abstract{Video Capsule Endoscopy (VCE) poses a challenging multi-label temporal classification problem, requiring simultaneous localization of 8 anatomical regions and detection of 9 pathological findings across tens of thousands of frames. We present GALAR-TemporalNet v2, a hierarchical temporal model that addresses three core challenges: extreme class imbalance, long-range temporal dependencies, and pathology--anatomy entanglement. Our architecture combines windowed self-attention for local modeling, a Dual-Graph GCN for global frame relationships, and Bidirectional Mamba for selective boundary context encoding. A novel anatomy prototype residual pathway decouples pathological deviation signals from normal organ appearance, and a frame-level GCN skip connection stabilizes training of visually confusable rare classes. The competition version, GALAR-TemporalNet, achieved an overall mAP@0.5 of 0.2644 and mAP@0.95 of 0.2353 on the RARE-VISION test set. Following the competition, the redesigned GALAR-TemporalNet v2---incorporating a restructured pathology branch, refined loss functions, and extended post-processing---improved these results to mAP@0.5 of 0.3409 and mAP@0.95 of 0.3333.}

\section{Motivation}\label{sec1}

\begin{figure}[!htb]
    \centering
    \includegraphics[width=0.88\linewidth]{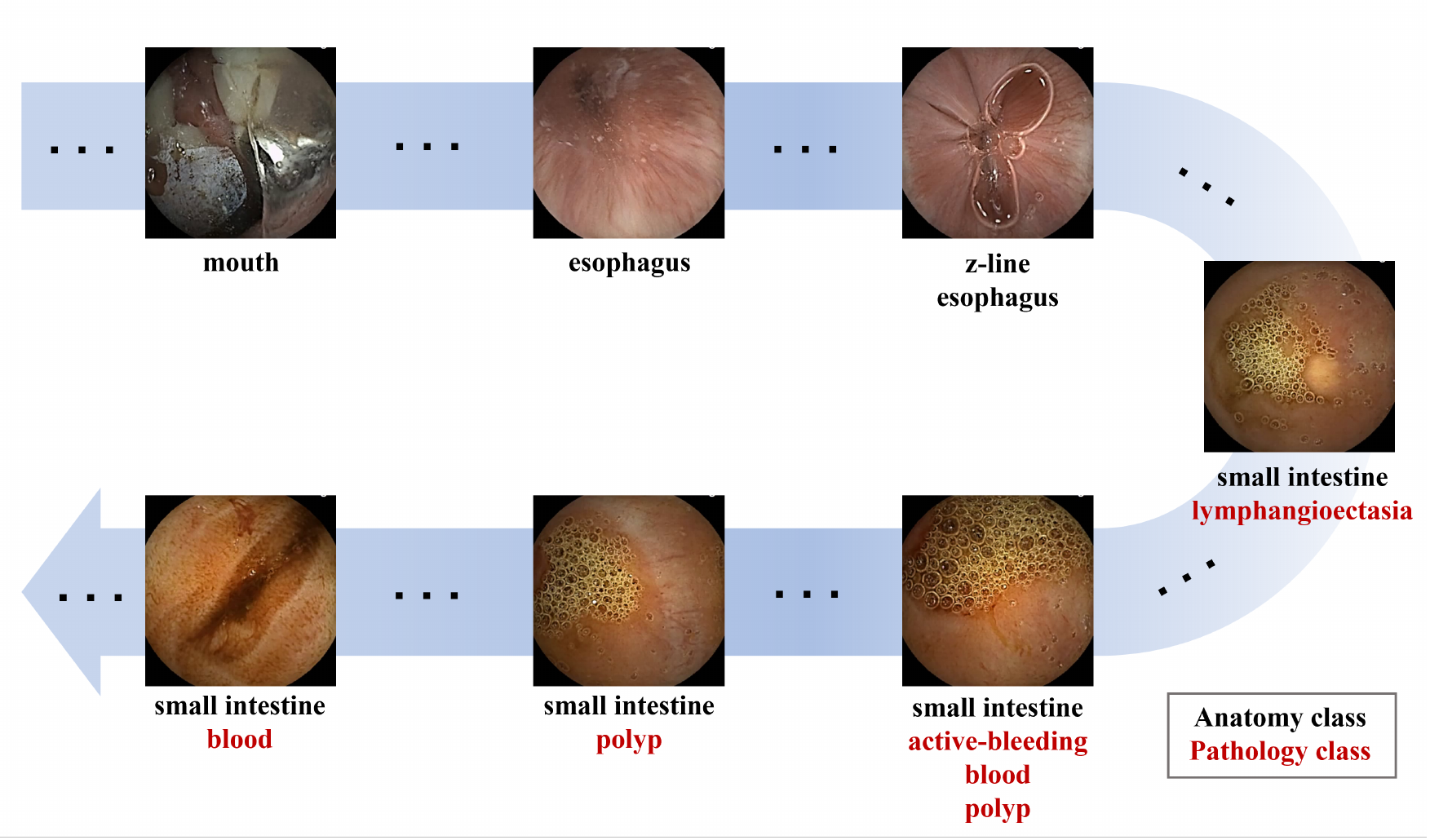}
    \vspace{-4mm}
    \caption{Representative labeling structure of the Galar VCE dataset.}
    \label{fig:dataset}
\end{figure}

Video Capsule Endoscopy (VCE) is a non-invasive diagnostic tool that captures the entire gastrointestinal tract as a continuous video stream, typically containing tens of thousands of frames. Automated analysis of the Galar dataset, a large multi-label VCE collection reflecting real-world clinical conditions, requires simultaneous classification of 17 categories: 8 anatomical location classes (mouth, esophagus, z-line, stomach, pylorus, small intestine, ileocecal valve, and colon) and 9 pathological findings (active bleeding, angiectasia, blood, erosion, erythema, hematin, lymphangioectasis, polyp, and ulcer). Representative examples of this labeling structure are shown in Figure~\ref{fig:dataset}.

Three core challenges motivate our design. First, \textbf{extreme class imbalance}: transitional anatomy classes such as the pylorus and z-line appear only briefly at organ-boundary moments, and many pathological findings are exceedingly rare relative to dominant normal frames. Standard cross-entropy training treats all frames equally and consequently fails to learn these underrepresented classes. Second, \textbf{long-range temporal structure}: the gastrointestinal tract follows a fixed anatomical sequence---mouth, esophagus, z-line, stomach, pylorus, small intestine, ileocecal valve, and colon---meaning that visually similar frames can carry different anatomical meaning depending on their global position in the video, a disambiguation that frame-level classifiers cannot perform. Third, \textbf{pathology--anatomy entanglement}: pathological findings manifest as deviations from the normal appearance of a specific organ, so treating anatomy and pathology as independent tasks ignores a natural hierarchical supervision signal. Our work addresses all three challenges through a unified temporal architecture.

\section{Methods}\label{sec2}

\begin{figure}[!htb]
    \centering
    \includegraphics[width=0.88\linewidth]{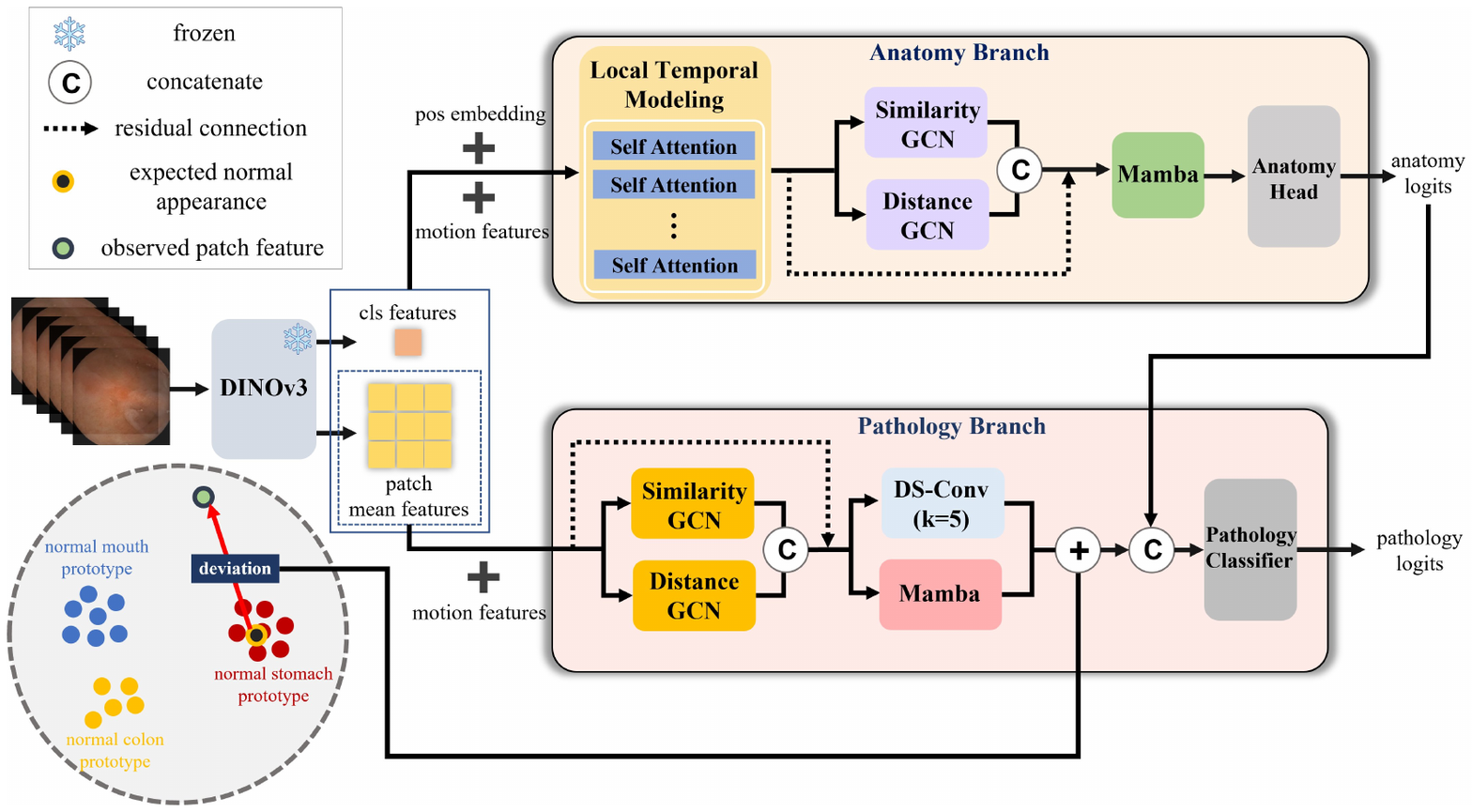}
    \vspace{-4mm}
    \caption{Overall architecture of GALAR-TemporalNet v2.}
    \label{fig:figure2}
\end{figure}

GALAR-TemporalNet v2 is a dual-branch temporal model that takes pre-extracted DINOv3 features as input and produces per-frame predictions for all 17 classes. The full pipeline is illustrated in Figure~\ref{fig:figure2}. The two branches are separated by design: the Anatomy Branch estimates where the camera is located in the gastrointestinal tract at each moment, while the Pathology Branch uses this anatomical context to detect deviations from normal organ appearance.

Input features are split into two streams: CLS tokens, which capture global frame-level semantics, and patch-mean features, which capture local texture information.

\textbf{Anatomy Branch.} The full 2048-dimensional feature is projected to 512 dimensions and enriched with a learnable positional embedding and a CLS motion signal, defined as the frame-to-frame difference of CLS tokens, which serves as an explicit boundary detector. Two layers of windowed self-attention then capture local temporal consistency. A Dual-Graph GCN follows, comprising a Similarity GCN that connects visually similar frames regardless of temporal distance and a Distance GCN that connects temporally nearby frames to enable smooth information propagation. The two outputs are concatenated, projected, and added to a saved residual. A Video GPS signal, defined as the scalar ratio $r = \text{window start frame} / \text{total frames}$ and encoded via an MLP, is then broadcast-added to all frames after the GCN, providing a global position prior without influencing the similarity adjacency computation. Finally, Bidirectional Mamba encodes selective forward and backward context, enabling detection of short-duration transition classes such as the pylorus and z-line. The anatomy classification head produces anatomy logits.

\textbf{Pathology Branch.} This branch constructs three complementary signals from raw patch features. Signal A, the deviation signal, computes the expected normal appearance for each frame by weighting per-anatomy healthy prototypes with detached anatomy predictions and then subtracts this estimate from the raw patch feature to isolate abnormal residuals; this signal bypasses the GCN and goes directly to fusion. Signal B, the motion signal, captures frame-to-frame patch differences as abrupt visual change indicators. Signal C, the content signal, combines the raw patch projection with the motion signal and serves as the input to a Pathology Dual-Graph GCN, which groups frames sharing similar lesion appearances while preserving each frame's own signal through a residual connection. Following the GCN, a depthwise-separable Conv1d (DS-Conv), comprising a depthwise convolution with kernel size 5 for per-channel local temporal mixing within a $\pm 2$-frame receptive field followed by a pointwise convolution for cross-channel feature recombination, captures brief localized lesion events that appear and disappear within a few frames. A subsequent Mamba module models longer-range temporal dependencies across the full sequence. The output of the depthwise-separable convolution, Mamba, and the pre-convolution GCN output are summed via a triple residual connection, ensuring that each module's contribution is preserved independently. The three pathology signals---deviation, motion, and temporally contextualized content after GCN, depthwise-separable convolution, and Mamba processing---are then concatenated and linearly projected to form the unified pathology representation, to which anatomy soft conditioning is added. Anatomy soft conditioning is introduced by passing detached anatomy logits through a small MLP and adding the result to the fused representation, allowing the model to learn location-aware pathology priors, for instance associating stomach regions with a higher likelihood of erosion or blood. The pathology classification head produces the final pathology logits.

\textbf{Inference.} Overlapping windows are averaged per frame, followed by median filtering with kernel size 5 to suppress noisy predictions. Anatomy probabilities are refined by Viterbi decoding to enforce the biological gastrointestinal traversal order and recover short-duration transitional classes. Co-occurrence gating then suppresses anatomically implausible pathology predictions using anatomy--pathology co-occurrence statistics computed from the training set. A minimum segment filter removes predicted segments shorter than 20 frames to eliminate spurious short-burst noise. Finally, an anatomy gap-filling step corrects short missing intervals in anatomy predictions.

\subsection{How was class imbalance handled?}

Class imbalance is addressed at three levels. At the loss level, Asymmetric Loss (ASL) aggressively suppresses easy-negative gradients, while class-specific positive-weight boosting for the pylorus, z-line, ileocecal valve, and esophagus, together with per-class pathology upweighting for erosion, blood, and angiectasia, compensates for rare classes. Transition-boundary weighting further emphasizes frames at anatomy and pathology event boundaries, and a monotonicity loss penalizes anatomical order violations.

At the sampling level, a weighted random sampler oversamples windows containing rare pathology frames. At the architecture level, the normal-patch prototype residual provides the pathology branch with a signal structurally biased toward deviations from healthy appearance rather than dominant normal texture.


\subsection{Post competition changes in the Methods}

Following the competition, the model was substantially redesigned based on analysis of failure cases and alignment with the official evaluation metric. The key changes are summarized below.

\textbf{Feature Representation.} The competition version used a shared 2048-dimensional feature for both branches. In the post-competition version, the input is split into CLS tokens for the anatomy branch and patch-mean features for the pathology branch, allowing each branch to operate on the feature type most suited to its task.

\textbf{Anatomy Branch.} The motion signal was refined to operate only on 1024-dimensional CLS tokens, reducing noise from patch-level motion. A residual connection was added around the Dual-Graph GCN to prevent information loss. The Video GPS injection point was moved from before the GCN to after the GCN, avoiding contamination of the cosine-similarity adjacency computation.

\textbf{Pathology Branch.} The single similarity GCN was replaced with a Dual-Graph GCN incorporating both similarity and temporal distance adjacency. A depthwise-separable Conv1d and a Mamba module were introduced after the GCN for short-event and long-range temporal modeling respectively, combined via triple residual. The three pathology signals are now explicitly fused via linear projection, and anatomy soft-conditioning was added as an additive prior. The prototype system was restructured from hidden-space prototypes to normal patch prototypes in raw feature
space.

\textbf{Loss Functions.} Pathology transition-boundary weighting was added to emphasize lesion onset and offset frames. Per-class pathology upweighting was introduced for erosion, blood, and angiectasia. A monotonicity loss was introduced to penalize predicted anatomical sequences that violate the known proximal-to-distal ordering. The prototype auxiliary loss from the competition version was removed.

\textbf{Validation and Post-Processing.} Model selection was realigned from frame-level mAP to temporal mAP@0.5 and mAP@0.95, consistent with the official evaluation metric. The inference pipeline was extended with co-occurrence gating to suppress anatomically implausible predictions, a minimum segment filter to remove spurious short-duration outputs, and an anatomy gap-filling step to correct missing prediction intervals.


\section{Results}\label{sec3}


\begin{table}[htbp]
    \centering
    \caption{Performance comparison before and after the competition.}
    \begin{tabular}{c|c|cc}
        \hline
        Video & Phase & mAP@0.5 & mAP@0.95 \\
        \hline
        
        \multirow{2}{*}{ukdd\_navi\_00051}
        & Before & 0.4153 & 0.4118 \\
        & After  & 0.4782 & 0.4706 \\
        \hline
        
        \multirow{2}{*}{ukdd\_navi\_00068}
        & Before & 0.2392 & 0.1766 \\
        & After  & 0.1912 & 0.1765 \\
        \hline
        
        \multirow{2}{*}{ukdd\_navi\_00076}
        & Before & 0.1388 & 0.1177 \\
        & After  & 0.3533 & 0.3529 \\
        \hline
        
        \multirow{2}{*}{Average (3 Videos)}
        & Before & 0.2644 & 0.2353 \\
        & After  & 0.3409 & 0.3333 \\
        \hline
        
        \multirow{2}{*}{Overall Score}
        & Before & \multicolumn{2}{c}{0.2499} \\
        & After  & \multicolumn{2}{c}{0.3371} \\
        
        \hline
    \end{tabular}
    \label{tab:competition_results}
\end{table}

\section{Discussion (if any)}\label{sec4}


Several directions remain open for future work. First, patch mean pooling discards spatial layout information within each frame; preserving per-token patch representations would enable the model to localize where within a frame an anomaly occurs. Second, temporal frame masking was used as a regularizer, but MAE-style spatial patch masking would provide a stronger pretext task by forcing reconstruction of local texture from surrounding context and yielding representations more directly suited to detecting small, localized lesions. Third, the current backbone, DINOv3, is used as a frozen feature extractor. An alternative direction would be to first train a frame-level classifier, for instance using a CNN backbone such as ResNet, to improve per-frame discriminability, and then apply temporal modeling on top of those representations. This two-stage approach may produce stronger per-frame features before temporal reasoning is introduced.

\section{Summary}\label{sec5}


%

Team Dubai Chewy Cookie utilized \textbf{GALAR-TemporalNet}, a hierarchical temporal model combining windowed self-attention, Dual-Graph GCN, and Bidirectional Mamba for multi-label VCE classification. Asymmetric Loss with class-specific positive-weight boosting, transition-boundary weighting, weighted random sampling, and a normal prototype residual at the architecture level were utilized to handle class imbalance in the dataset. A key finding is that explicitly modeling pathology as deviation from healthy anatomy prototypes provides a more principled signal than treating anatomy and pathology as independent tasks. The team achieved an overall mAP@0.5 of 0.2644 and an overall mAP@0.95 of 0.2353.

After the competition was over, we improved on \textbf{GALAR-TemporalNet v2}. The improvements included separating CLS and patch features into dedicated branches; redesigning the pathology branch with Dual-Graph GCN, depthwise-separable Conv1d, Mamba, and anatomy soft conditioning; introducing monotonicity loss and per-class pathology upweighting; aligning model selection with temporal mAP; and extending post-processing with co-occurrence gating, minimum segment filtering, and anatomy gap filling. Aligning the training objective with the official temporal mAP metric---rather than frame-level mAP---proved to be a critical factor in the performance gain. The team achieved an improved overall mAP@0.5 of 0.3409 and an overall mAP@0.95 of 0.3333.

\section{Acknowledgments}\label{sec6}

As participants in the ICPR 2026 RARE-VISION Competition, we fully complied with the competition's rules as outlined in \cite{Lawniczak2025, rarevision2026github}. Our AI model development was based exclusively on the datasets in the competition \cite{LeFloch2025figshareplus}. The mAP values were reported using the test dataset \cite{rarevision_testdata_2026}, and sanity checker \cite{manasapp} released in the competition.

\bibliographystyle{unsrtnat}
\bibliography{sample}

\end{document}